\documentclass[runningheads]{llncs}
\usepackage{graphicx}

\usepackage{tikz}
\usepackage{comment}
\usepackage{amsmath,amssymb} 
\usepackage{color}

\usepackage[accsupp]{axessibility}  

\usepackage{comment,wrapfig,caption,subcaption}
\usepackage{booktabs,tabulary,pifont,multirow,xspace,xcolor,makecell}
\usepackage{dsfont,fixmath,mathtools,nicefrac,bm,colortbl,mmstyles,marvosym}

\newcommand{\methodname}{OV-DETR\xspace}
\definecolor{ForestGreen}{RGB}{34,139,34}
\definecolor{Cerulean}{rgb}{0.0, 0.48, 0.65}
\definecolor{LavenderBlush}{rgb}{1.0, 0.94, 0.96}
\newcommand{\rowNumber}[1]{\textcolor{Cerulean}{#1}}
\newcommand{\better}[1]{\textcolor{ForestGreen}{(+#1)}}
\newcommand{\cmark}{\ding{51}}%
\newcommand{\xmark}{\ding{55}}%

\definecolor{citecolor}{HTML}{0071bc}

\makeatletter
\renewcommand\paragraph{
  \@startsection{paragraph} 
  {4} 
  {\z@} 
  {.5em \@plus1ex \@minus.2ex} 
  {-1.5em} 
  {\normalfont\normalsize\bfseries} 
}
\makeatother

\begin{document}
\pagestyle{headings}
\mainmatter
\def\ECCVSubNumber{}  

\title{Open-Vocabulary DETR \\ with Conditional Matching}

\titlerunning{Open-Vocabulary DETR with Conditional Matching}
%
\author{
{
Yuhang Zang$^{1}$, Wei Li$^{1}$, Kaiyang Zhou$^{1}$, Chen Huang$^{2}$, Chen Change Loy$^{1}$\textsuperscript{\Letter}} 
\\
{
${^1}$S-Lab, Nanyang Technological University ~~${^2}$Carnegie Mellon University
}
\\
{\tt\small $\{$zang0012, wei.l, kaiyang.zhou, ccloy$\}$@ntu.edu.sg ~~chen-huang@apple.com}
}
\authorrunning{
Y. Zang, W. Li, K. Zhou, C. Huang, C.C.Loy
}
%
\institute{}
\maketitle
\begin{abstract}
Open-vocabulary object detection, which is concerned with the problem of detecting novel objects guided by natural language, has gained increasing attention from the community. Ideally, we would like to extend an open-vocabulary detector such that it can produce bounding box predictions based on user inputs in form of either natural language or exemplar image. This offers great flexibility and user experience for human-computer interaction. To this end, we propose a novel open-vocabulary detector based on DETR---hence the name OV-DETR---which, once trained, \emph{can detect any object given its class name or an exemplar image}. The biggest challenge of turning DETR into an open-vocabulary detector is that it is impossible to calculate the classification cost matrix of novel classes without access to their labeled images.
To overcome this challenge, we formulate the learning objective as a binary matching one between input queries (class name or exemplar image) and the corresponding objects, which learns useful correspondence to generalize to unseen queries during testing. For training, we choose to condition the Transformer decoder on the input embeddings obtained from a pre-trained vision-language model like CLIP, in order to enable matching for both text and image queries.
With extensive experiments on LVIS and COCO datasets, we demonstrate that our OV-DETR---\emph{the first end-to-end Transformer-based open-vocabulary detector}---achieves non-trivial improvements over current state of the arts. Code is available at \url{https://github.com/yuhangzang/OV-DETR}.
\end{abstract}
\section{Introduction}
\begin{figure}[!t]
    \centering
    \includegraphics[page=1, width=\linewidth]{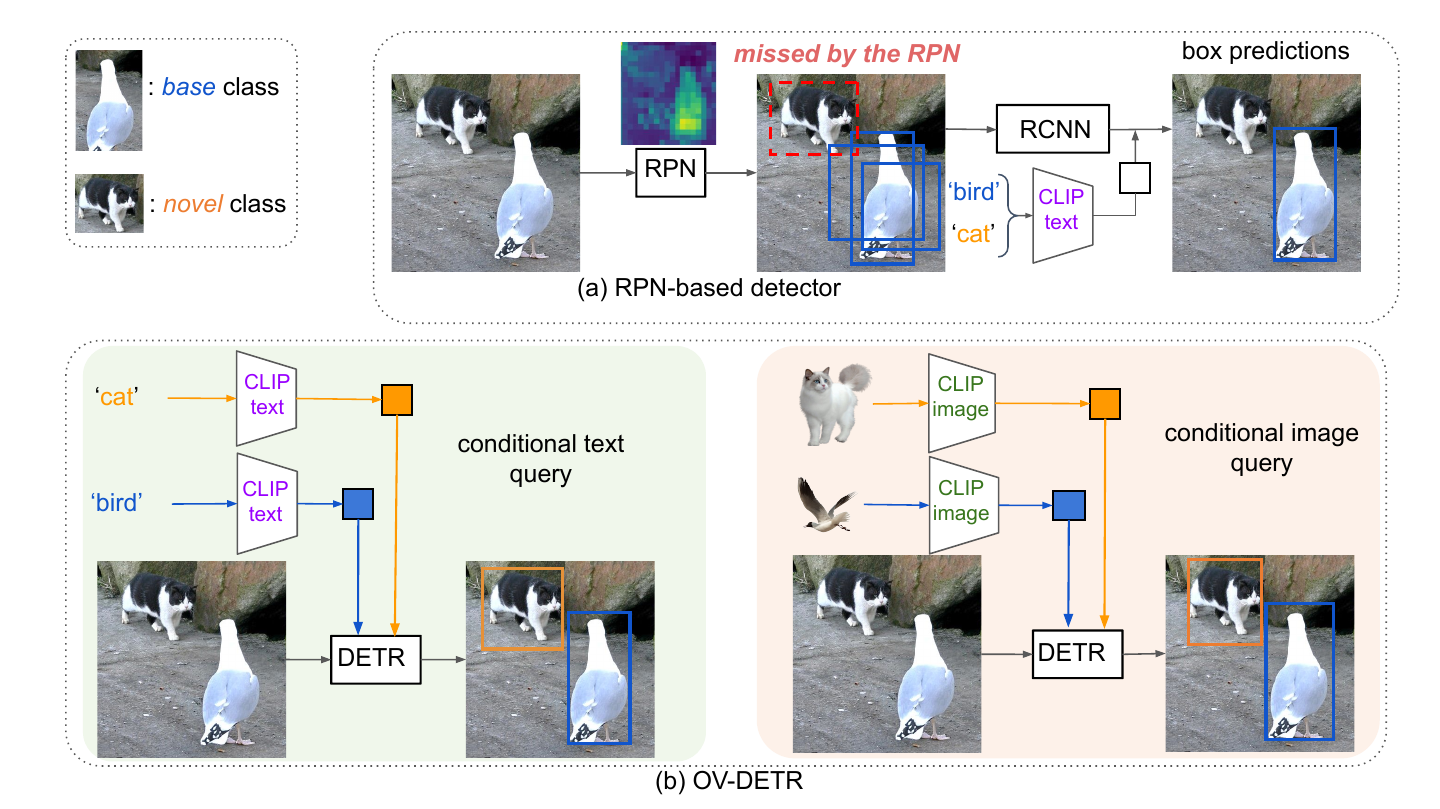}
    \caption{
    \textbf{Comparison between a RPN-based detector and our Open-Vocabulary Transformer-based detector (OV-DETR) using conditional queries}. The RPN trained on closed-set object classes is easy to ignore novel classes (\eg,~the ``cat'' region receives little response). Hence the cats in this example are largely missed with few to no proposals. By contrast, our OV-DETR is trained to perform matching between a conditional query and its corresponding box, which helps to learn correspondence that can generalize to queries from unseen classes. Note we can take input queries in the form of either text (class name) or exemplar images, which offers greater flexibility for open-vocabulary object detection.
    }
    \label{fig:teaser}
\end{figure}

Object detection, a fundamental computer vision task aiming to localize objects with tight bounding boxes in images, has been significantly advanced in the last decade thanks to the emergence of deep learning~\cite{papageorgiou2000trainable,szegedy2013deep,felzenszwalb2008discriminatively,ren2015faster,he2017mask}. However, most object detection algorithms are unscalable in terms of the vocabulary size, \ie, they are limited to a fixed set of object categories defined in detection datasets~\cite{everingham2010pascal,lin2014microsoft}. For example, an object detector trained on COCO~\cite{lin2014microsoft} can only detect 80 classes and is unable to handle new classes beyond the training ones.

A straightforward approach to detecting novel classes is to collect and add their training images to the original dataset, and then re-train or fine-tune the detection model. This is, however, both impractical and inefficient due to the large cost of data collection and model training. In the detection literature, generalization from base to novel classes has been studied as a zero-shot detection problem~\cite{bansal2018zero} where zero-shot learning techniques like word embedding projection~\cite{frome2013devise} are widely used.

Recently, open-vocabulary detection, a new formulation that leverages large pre-trained language models, has gained increasing attention from the community~\cite{zareian2021open,gu2021open}. The central idea in existing works is to align detector's features with embedding provided by models pre-trained on large scale image-text pairs like CLIP~\cite{radford2021learning}~(see Fig~\ref{fig:teaser}~(a)). This way, we can use an aligned classifier to recognize novel classes only from their descriptive texts.

A major problem with existing open-vocabulary detectors~\cite{zareian2021open,gu2021open} is that they rely on region proposals that are often not reliable to cover all novel classes in an image due to the lack of training data, see Fig.~\ref{fig:teaser}(a). This problem has also been identified by a recent study~\cite{kim2021learning}, which suggests the binary nature of the region proposal network (RPN) could easily lead to overfitting to seen classes (thus fail to generalize to novel classes).

In this paper, we propose to train end-to-end an open-vocabulary detector under the Transformer framework, aiming to enhance its novel class generalization without using an intermediate RPN. To this end, we propose a novel open-vocabulary detector based on DETR~\cite{carion2020end}---hence the name OV-DETR---which is trained to \emph{detect any object given its class name or an exemplar image}. This would offer greater flexibility than conventional open-vocabulary detection from natural language only.

Despite the simplicity of end-to-end DETR training, turning it into an open-vocabulary detector is non-trivial. The biggest challenge is the inability to calculate the classification cost for novel classes without their training labels. To overcome the challenge, we re-formulate the learning objective as binary matching between input queries (class name or exemplar image) and the corresponding objects. Such a matching loss over diverse training pairs allows to learn useful correspondence that can generalize to unseen queries during testing. For training, we extend the Transformer decoder of DETR to take conditional input queries. Specifically, we condition the Transformer decoder on the query embeddings obtained from a pre-trained vision-language model CLIP~\cite{radford2021learning}, in order to perform conditional matching for either text or image queries. Fig.~\ref{fig:teaser} shows this high-level idea, which proves better at detecting novel classes than RPN-based closed-set detectors.

We conduct comprehensive experiments on two challenging
open-vocabulary object detection datasets, and show consistent improvements in performance. Concretely, our \methodname method achieves $17.4$ mask mAP of novel classes on the open-vocabulary LVIS dataset~\cite{gu2021open} and $29.4$ box mAP of novel classes on open-vocabulary COCO dataset~\cite{zareian2021open}, surpassing SOTA methods by $1.3$ and $1.8$ mAP, respectively.
\section{Related Work}

\noindent \textbf{Open-Vocabulary Object Detection} leverages the recent advances in large pre-trained language models~\cite{zareian2021open,gu2021open} to incorporate the open-vocabulary information into object detectors. OVR-CNN~\cite{zareian2021open} first uses BERT~\cite{devlin2018bert} to pre-train the Faster R-CNN detector~\cite{ren2015faster} on image-caption pairs and then fine-tunes the model on downstream detection datasets. ViLD~\cite{gu2021open} adopts a distillation-based approach that aligns the image feature extractor of Mask R-CNN~\cite{he2017mask} with the image and text encoder of CLIP~\cite{radford2021learning} so the CLIP can be used to synthesize the classification weights for any novel class. 
The prompt tuning techniques~\cite{zhou2022coop,zhou2022cocoop,zhang2022neural} for the pre-trained vision-language model have also been applied for open-vocabulary detectors, like DetPro~\cite{du2022learning}.
Our approach differs from these works in that we train a Transformer-based detector end-to-end, with a novel framework of conditional matching.

\noindent \textbf{Zero-Shot Object Detection}
is also concerned with the problem of detecting novel classes~\cite{bansal2018zero,li2019zero,zhu2020don,rahman2021improved,zhao2020gtnet}. However, this setting is less practical due to the harsh constraint of limiting access to resources relevant to unseen classes~\cite{zareian2021open}. A common approach to zero-shot detection is to employ word embeddings like GloVe~\cite{pennington2014glove} as the classifier weights~\cite{bansal2018zero}. Other works have found that using external resources like textual descriptions can help improve the generalization of classifier embeddings~\cite{li2019zero,rahman2021improved}. Alternatively, Zhao \etal.~\cite{zhao2020gtnet} used Generative Adversarial Network (GAN)~\cite{goodfellow2014generative} to generate feature representations of novel classes. While Zhu \etal.~\cite{zhu2020don} synthesized unseen classes using a data augmentation strategy.

\noindent \textbf{Visual Grounding}
is another relevant research area where the problem is to ground a target object in one image using natural language input~\cite{deng2018visual,chen2017query}. Different from open-vocabulary detection that aims to identify all target objects in an image, the visual grounding methods typically involve a particular single object, hence cannot be directly applied to generic object detection. There is a relevant visual grounding method though, which is called MDETR~\cite{kamath2021mdetr}. This method similarly trains DETR along with a given language model so as to link the output tokens of DETR with specific words. MDETR also adopts a conditional framework, where the visual and textual features are combined to be fed to the Transformer encoder and decoder. However, the MDETR method is not applicable to open-vocabulary detection because it is unable to calculate the cost matrix for novel classes under the classification framework. Our OV-DETR bypasses this challenge by using a conditional matching framework instead.

\noindent \textbf{Object Detection with Transformers.}
The pioneer DETR approach~\cite{carion2020end} greatly simplifies the detection pipeline by casting detection as a set-to-set matching problem. Several follow-up methods have been developed to improve performance and training efficiency. Deformable DETR~\cite{zhu2020deformable} features a deformable attention module, which samples sparse pixel locations for computing attention, and further mitigates the slow convergence issue with a multi-scale scheme. SMCA~\cite{gao2021fast} accelerates training convergence with a location-aware co-attention mechanism. Conditional DETR~\cite{meng2021conditional} also addresses the slow convergence issue, but with conditional spatial queries learned from reference points and the decoder embeddings. Our work \emph{for the first time} extends DETR to the open-vocabulary domain by casting open-vocabulary detection as a conditional matching problem, and achieves non-trivial improvements over current SOTA.

\section{Open-Vocabulary DETR}

Our goal is to design a simple yet effective open-vocabulary object detector that can detect objects described by arbitrary text inputs or exemplar images. We build on the success of DETR~\cite{carion2020end} that casts object detection as an end-to-end set matching problem (among closed classes), thus eliminating the need of hand-crafted components like anchor generation and non-maximum suppression.
This pipeline makes it appealing to act as a suitable framework to build our end-to-end open-vocabulary object detector.

However, it is non-trivial to retrofit a standard DETR with closed-set matching to an open-vocabulary detector that requires matching against unseen classes. One intuitive approach for such open-set matching is to learn a class-agnostic module (\eg, ViLD~\cite{gu2021open}) to handle all classes. This is, however, still unable to match for those open-vocabulary classes that come with no labeled images. Here we provide a new perspective on the matching task in DETR, which leads us to reformulate the fixed set-matching objective into a conditional binary matching one between conditional inputs (text or image queries) and detection outputs.

\begin{figure*}[!t]
\centering
\includegraphics[width=0.99\textwidth]{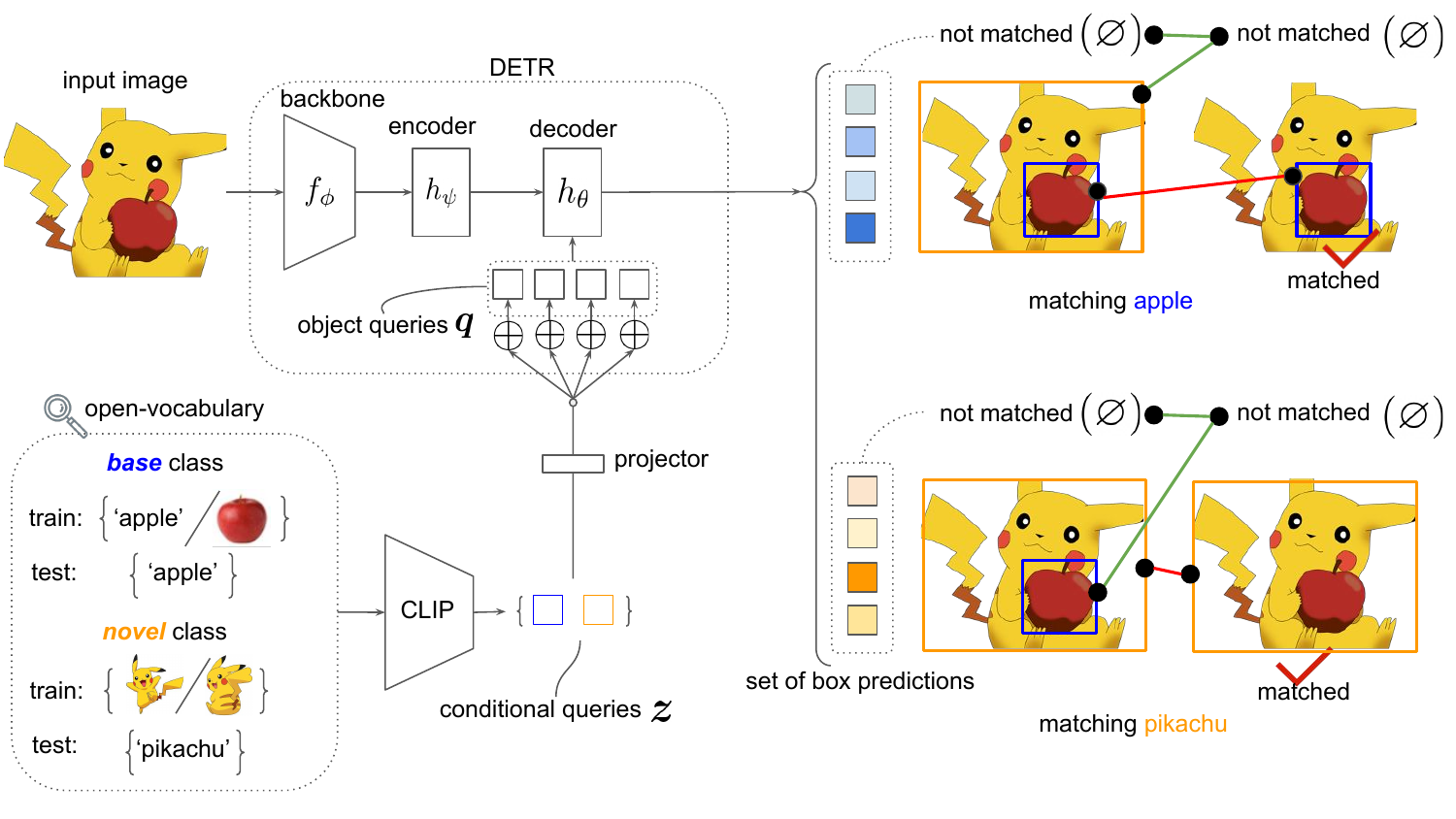}
\caption{
\textbf{Overview of \methodname}. Unlike the standard DETR, our method does not separate `objects' from `non-objects' for a closed set of classes. Instead, \methodname performs open-vocabulary detection by measuring the matchability (`matched' vs. `not matched') between some conditional inputs (text or exemplar image embeddings from CLIP) and detection results. We show such pipeline is flexible to detect open-vocabulary classes with arbitrary text or image inputs. 
}
\label{fig:pipeline}
\end{figure*}

An overview of our Open-Vocabulary DETR is shown in Fig. \ref{fig:pipeline}. At high level, DETR first takes query embeddings (text or image) as conditional inputs obtained from a pre-trained CLIP~\cite{radford2021learning} model, and then a binary matching loss is imposed against the detection result to measure their matchability. In the following, we will revisit the closed-set matching process in standard DETR in Section \ref{subsec:revist_matching}. We then describe how to perform conditional binary matching in our OV-DETR in Section \ref{subsec:conditional_matching}.

\subsection{Revisiting Closed-Set Matching in DETR}
\label{subsec:revist_matching}
For input image $\vx$, a standard DETR infers $N$ object predictions $\hat{\vy}$ where $N$ is determined by the fixed size of object queries $\vq$ that serve as learnable positional encodings. One single pass of the DETR pipeline consists of two main steps: (i) set prediction, and (ii) optimal bipartite matching.

\noindent \textbf{Set Prediction.} Given an input image $\vx$, the global context representations $\vc$ is first extracted by a CNN backbone $f_\phi$ and then a Transformer encoder $h_\psi$:
\begin{equation}
\label{eq:detr_feat}
\vc = h_\psi(f_\phi(\vx)),
\end{equation}
where the output $\vc$ denotes a sequence of feature embeddings of $\vq$. Taking the context feature $\vc$ and object queries $\vq$ as inputs,
the Transformer decoder $h_\theta$ (with prediction heads) then produce the set prediction $\hat{\vy} = \{\hat{\vy}_i\}_{i=1}^{N}$ :
\begin{equation}
\label{eq:detr_set}
\hat{\vy} = h_\theta(\vc, \vq),
\end{equation}
where $\hat{\vy}$ contains both bounding box predictions $\hat{\vb}$ and class predictions $\hat{\vp}$ for a closed-set of training classes.

\noindent \textbf{Optimal Bipartite Matching} is to find the best match between the set of $N$ predictions $\hat{\vy}$ and the set of ground truth objects $\vy = \{\vy_i\}_{i=1}^{M}$ (including no object $\varnothing$). Specifically, one needs to search a permutation of $N$ elements $\sigma \in \mathfrak{S}_N$ that has the lowest matching cost: 
\begin{equation}
\label{eq:set_matching}
\hat{\sigma} = \argmin_{\sigma \in \mathfrak{S}_N} \sum_i^N \mathcal{L}_{cost}(\vy_i,\hat{\vy}_{\sigma(i)}),
\end{equation}
where $\mathcal{L}_{cost}(\vy_i, \hat{\vy}_{\sigma(i)})$ is a pair-wise \textit{matching cost} between ground truth $\vy_i$ and the prediction $\hat{\vy}_{\sigma(i)}$ with index $\sigma(i)$. Note $\mathcal{L}_{cost}$ is comprised of the losses for both class prediction $\mathcal{L}_{\text{cls}}(\hat{\vp}, \vp)$ and bounding box localization $\mathcal{L}_{\text{box}}(\hat{\vb}, \vb)$. The whole bipartite matching process produces \textit{one-to-one} label assignments, where each prediction $\hat{\vy}_{i}$ is assigned to a ground-truth annotation $\vy_{j}$ or $\varnothing$~(no object). The optimal assignment can be efficiently found by the Hungarian algorithm~\cite{kuhn1955hungarian}.

\begin{figure*}[t]
\centering
\includegraphics[width=0.96\textwidth]{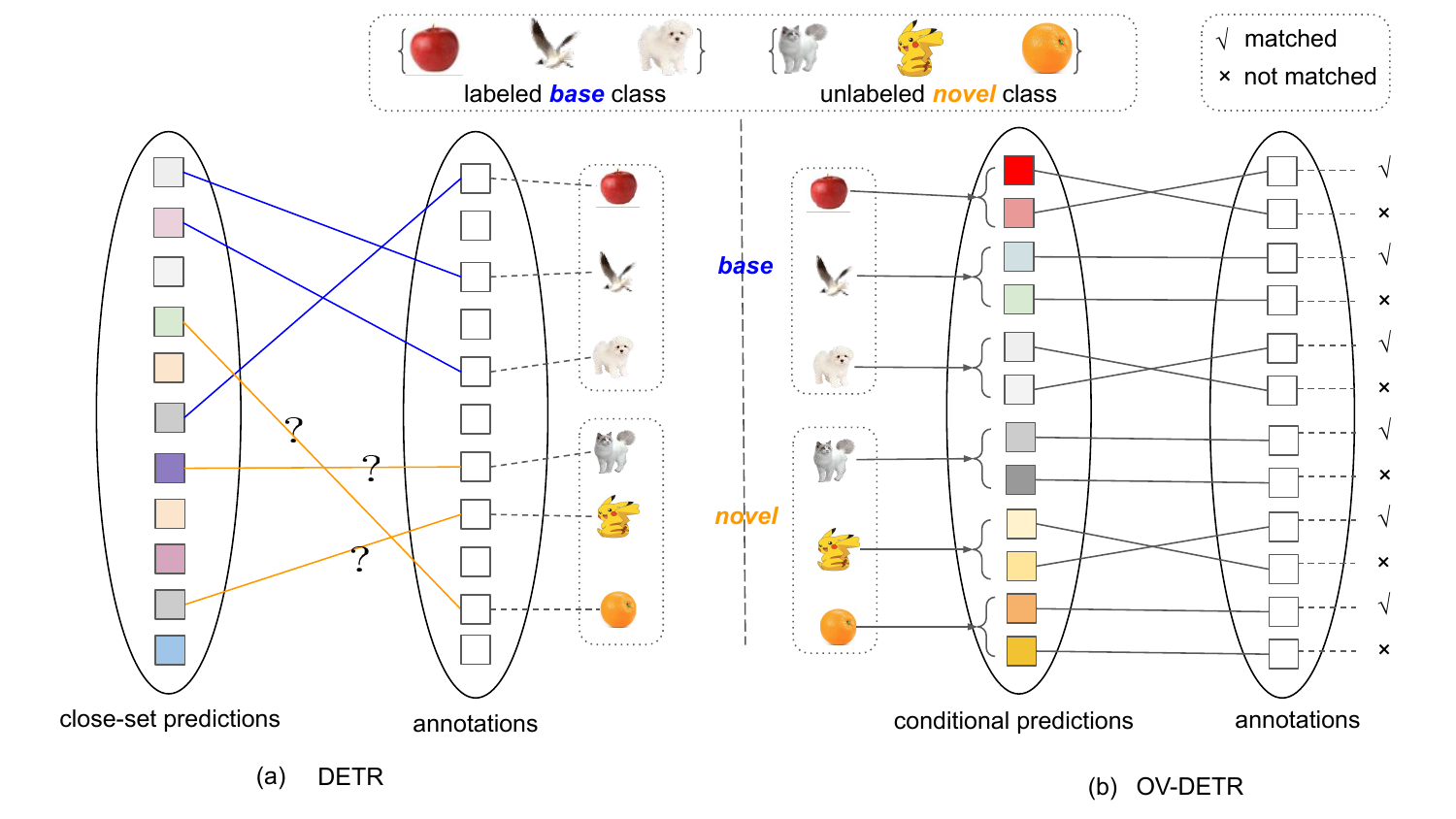}
\caption{
\textbf{Comparing the label assignment mechanisms of DETR and our OV-DETR}.
\textbf{(a)} In the original DETR, the set-to-set prediction is conducted via bipartite matching between predictions and closed-set annotations, in which a cost matrix in respect of the queries and categories. Due to the absence of class label annotations for \emph{novel} classes, computing such a class-specific cost matrix is impossible.
\textbf{(b)} On the contrary, our OV-DETR casts the open-vocabulary detection as a conditional matching process and formulate a binary matching problem that computes a class-agnostic matching cost matrix for conditional inputs.
}
\label{fig:label_assignment}
\end{figure*}

\noindent \textbf{Challenge.} As mentioned above, the bipartite matching method cannot be directly applied to an open-vocabulary setting that contains both \emph{base} and \emph{novel} classes. The reason is that computing the matching cost in Eq.~(\ref{eq:set_matching}) requires access of the label information, which is unavailable for \emph{novel} classes.
We can follow previous works~\cite{gu2021open,xie2021zsd,du2022learning} to generate class-agnostic object proposals that may cover the \emph{novel} classes, but we do not know the ground-truth classification labels of these proposals.
As a result, the predictions for the $N$ object queries cannot generalize to novel classes due to the lack of training labels for them.
As shown in Fig.~\ref{fig:label_assignment}~(a), bipartite matching can only be performed for base classes with available training labels.

\subsection{Conditional Matching for Open-Vocabulary Detection}\label{sec:conditional_matching}
\label{subsec:conditional_matching}
To enable DETR to go beyond closed-set classification and perform open-vocabulary detection, we equip the Transformer decoder with conditional inputs and reformulate the learning objective as binary matching problem.

\noindent \textbf{Conditional Inputs.} Given an object detection dataset with standard annotations for all the training (\emph{base}) classes, we need to convert those annotations to conditional inputs to facilitate our new training paradigm. Specifically, for each ground-truth annotation with bounding box $\vb_i$ and class label name $\vy^{\text{class}}_i$, we use the CLIP model~\cite{radford2021learning} to generate their corresponding image embedding $\vz_{i}^{\text{image}}$ and text embedding $\vz_{i}^{\text{text}}$:
\begin{equation}
    \begin{aligned}
        \vz_{i}^{\text{image}}  &= \text{CLIP}_{image}(\vx, \vb_i), \\
        \vz_{i}^{\text{text}} &= \text{CLIP}_{text}(\vy^{\text{class}}_i).
    \end{aligned}
\end{equation}
Such image and text embeddings are already well-aligned by the CLIP model. Therefore, we can choose either of them as input queries to condition the DETR's decoder and train to match the corresponding objects. Once training is done, we can then take arbitrary input queries during testing to perform open-vocabulary detection. To ensure equal training conditioned on image and text queries, we randomly select $\vz_{i}^{\text{text}}$ or $\vz_{i}^{\text{image}}$ with probability $\xi=0.5$
as conditional inputs. Moreover, we follow previous works~\cite{gu2021open,xie2021zsd,du2022learning} to generate additional object proposals for \emph{novel} classes to enrich our training data. We only extract image embeddings $\vz_{i}^{\text{image}}$ for such novel-class proposals as conditional inputs, since their class names are unavailable to extract text embeddings. Please refer to supplementary materials for more details.

\begin{figure*}[t]
\centering
\includegraphics[width=1\textwidth]{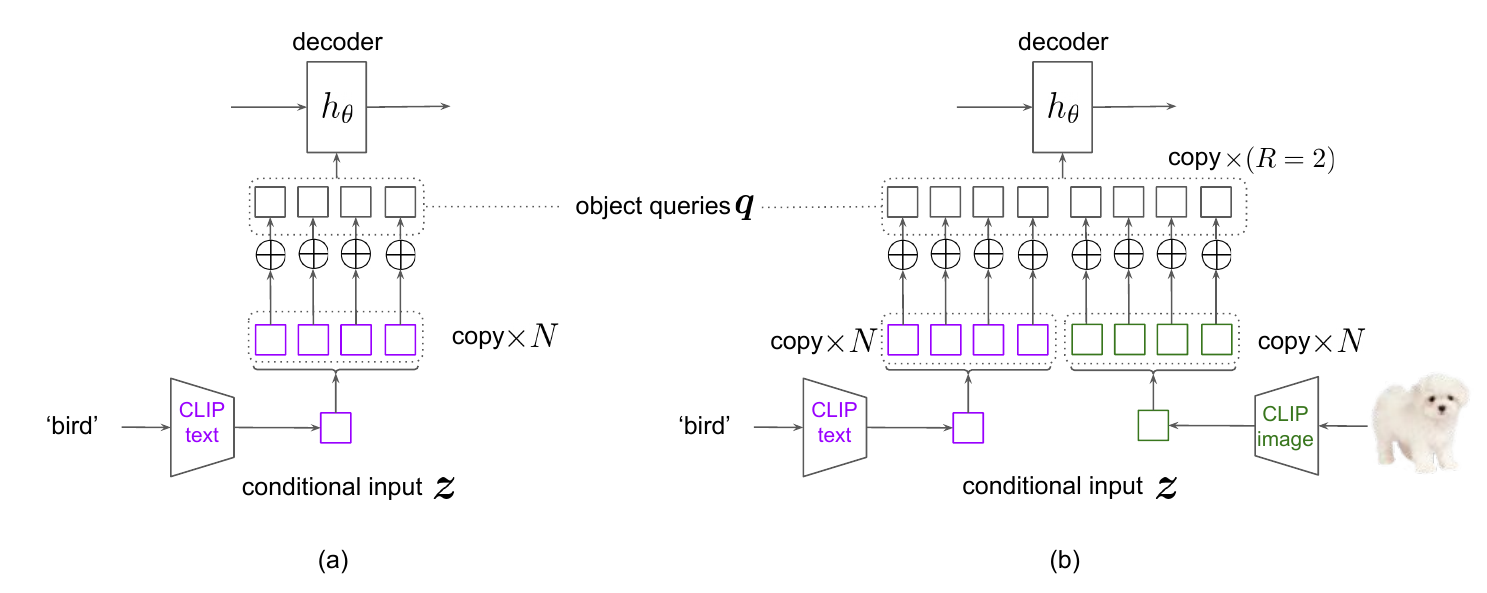}
\caption{
DETR decoder with \textbf{(a)} single conditional input or \textbf{(b)} multiple conditional inputs in parallel.
}
\label{fig:conditional_query}
\end{figure*}
\noindent \textbf{Conditional Matching.} Our core training objective is to measure the matchability between the conditional input embeddings and detection results. In order to perform such conditional matching, we start with a fully-connected layer $\mathds{F}_{\text{proj}}$ to project the conditional input embeddings ($\vz_{i}^{text}$ or $\vz_{i}^{image}$) to have the same dimension as $\vq$. Then the input to the DETR decoder $\vq^{\prime}$ is given by:
\begin{equation}
    \vq^{\prime} = \vq \oplus \mathds{F}_{\text{proj}}(\vz_{i}^{\text{mod}}), \:\: \text{mod}\in \{ \text{text},\text{image} \},
\label{eq:conditioning}    
\end{equation}
where we use a simple addition operation $\oplus$ to convert the \textit{class-agnostic} object queries $\vq$ into \textit{class-specific} $\vq^{\prime}$ informed by $\mathds{F}_{\text{proj}}(\vz_{i}^{\text{mod}})$.

In practice, adding the conditional input embeddings $\vz$ to only one object query will lead to a very limited coverage of the target objects that may appear many times in the image. Indeed, in existing object detection datasets, there are typically multiple object instances in each image from the same or different classes.
To enrich the training signal for our conditional matching, we copy the object queries $\vq$ for $R$ times, and the conditional inputs ($\vz_{i}^{\text{text}}$ or $\vz_{i}^{\text{image}}$) for $N$ times before performing the conditioning in Eq.~(\ref{eq:conditioning}). As a result, we obtain a total of $N \times R$ queries for matching during each forward pass, as shown in Fig.~\ref{fig:conditional_query}~(b). Experiments in the supplementary material will validate the importance of such ``feature cloning'' and also show how we determine $N$ and $R$ based on the performance-memory trade-off. Note for the final conditioning process, we further add an attention mask to ensure the independence between different query copies, as is similarly done in~\cite{dai2021up}.

Given the conditioned query features $\vq^{\prime}$, our binary matching loss for label assignment is given as:
\begin{equation}
    \label{eq:match_cost}
    \mathcal{L}_{\text{cost}}\left(\vy, \hat{\vy}_{\sigma}\right) = \mathcal{L}_{\mathrm{match}}\left(\vp, \hat{\vp}_{\sigma}\right) + \mathcal{L}_{\mathrm{box}}\left(\vb, \hat{\vb}_{\sigma}\right),
\end{equation}
where $\mathcal{L}_{\mathrm{match}}\left(\vp, \hat{\vp}_{\sigma}\right)$ denotes a new matching loss that replaces the classification loss $\mathcal{L}_{\mathrm{cls}}\left(\vp, \hat{\vp}_{\sigma}\right)$ in Eq.~(\ref{eq:set_matching}).
Here in our case, $\vp$ is a 1-dimensional sigmoid probability vector that characterizes the matchability (`matched' vs. `not matched'), and $\mathcal{L}_{\mathrm{match}}$ is simply implemented by a Focal loss~\cite{lin2017focal} $\mathcal{L}_{\mathrm{Focal}}$ between predicted $\hat{\vp}_{\sigma}$ and groud-truth $\vp$.
For instance, with the `bird' query as input, our matching loss should allow us to match all the bird instances in one image, while tagging instances from other classes as `not matched'.

\subsection{Optimization}
After optimizing Eq.~(\ref{eq:match_cost}), we obtain the optimized label assignments $\sigma$ for different object queries. This process produces a set of detected objects with assigned box coordinates $\hat{\vb}$ and 2-dim matching probability $\hat{\vp}$ that we will use to compute our final loss function for modeling training. We further attach an embedding reconstruction head to the model, which learns to predict embedding $\ve$ to be able to reconstruct each conditional input embedding $\vz^\text{text}$ or $\vz^\text{image}$:
\begin{equation}
    \mathcal{L}_{\text{embed}}(\ve, \vz) = \left\| \ve - \vz^{\text{mod}} \right\|_{1}, \:\: \text{mod}\in \{ \text{text},\text{image} \}.
\end{equation}
Supplementary materials validate the effectiveness of $\mathcal{L}_{\text{embed}}$.

Our final loss for model training combines $\mathcal{L}_{\text{embed}}$ with bounding box losses $\mathcal{L}_{\text{match}}(\vp, \hat{\vp})$ and $\mathcal{L}_{\text{box}}(\vb, \hat{\vb})$ again:
\begin{equation}
    \begin{aligned}
        \mathcal{L}_{\text{loss}}(\vy, \hat{\vy})&=\mathcal{L}_{\text{match}}(\vp, \hat{\vp})+\mathcal{L}_{\text{box}}(\vb, \hat{\vb})+\mathcal{L}_{\text{embed}}(\ve, \vz) \\
    &=\lambda_{L_{\text{Focal}}}\mathcal{L}_{\text{Focal}}+\lambda_{L_{\text{L1}}}\mathcal{L}_{\text{L1}}+\lambda_{L_{\text{GIoU}}}\mathcal{L}_{\text{GIoU}}+\lambda_{L_{\text{embed}}}\mathcal{L}_{\text{embed}},
    \end{aligned}
\end{equation}
where $\mathcal{L}_{\text{box}}$ consists of the L1 loss and the generalized IoU (GIoU)~\cite{rezatofighi2019generalized} loss for boxes, while $\lambda_{L_{\text{Focal}}}$, $\lambda_{L_{\text{L1}}}$, $\lambda_{L_{\text{Giou}}}$ and $\lambda_{L_{\text{embed}}}$ are the  weighting parameters.

\subsection{Inference} \label{sec:inference}
During testing, for each image, we send the text embedding $\vz^{\text{text}}$ of all the \textit{base}+\textit{novel} classes to the model and merge the results by selecting the top $k$ predictions with highest prediction scores.
We follow the prior work~\cite{gu2021open} to use $k = 100$ for COCO dataset and $k = 300$ for LVIS dataset.
To obtain the context representation $\vc$ in Eq. (\ref{eq:detr_feat}), we forward the input image through the CNN backbone $f_\phi$ and Transformer encoder $h_\psi$. Note $\vc$ is computed only once and shared for all conditional inputs for efficiency. Then the conditioned object queries from different classes are sent to the Transformer decoder in parallel. In practice, we copy the object queries for $R$ times as shown in Fig.~\ref{fig:conditional_query}~(b).
\section{Experiments}

\noindent \textbf{Datasets.}
We evaluate our approach on two standard open-vocabulary detection benchmarks modified from LVIS~\cite{gupta2019lvis} and COCO~\cite{lin2014microsoft} respectively. LVIS~\cite{gupta2019lvis} contains 100K images with 1,203 classes. The classes are divided into three groups, namely frequent, common and rare, based on the number of training images. Following ViLD~\cite{gu2021open}, we treat 337 rare classes as \emph{novel} classes and use only the frequent and common classes for training. The COCO~\cite{lin2014microsoft} dataset is a widely-used benchmark for object detection, which consists of $80$ classes. Following OVR-CNN~\cite{zareian2021open}, we divide the classes in COCO into 48 \emph{base} categories and 17 \emph{novel} categories, while removing 15 categories without a synset in the WordNet hierarchy. The training set is the same as the full COCO but only images containing at least one \emph{base} class are used. We refer to these two benchmarks as OV-LVIS and OV-COCO hereafter.

\noindent \textbf{Evaluation Metrics.}
For OV-LVIS, we report the mask mAP for rare, common and frequent classes, denoted by $\rm{AP^{m}_{r}}$, $\rm{AP^{m}_{c}}$ and $\rm{AP^{m}_{f}}$. The rare classes are treated as \emph{novel} classes ($\rm{AP^{m}_{novel}}$). The symbol $\rm{AP^{m}}$ denotes to the mAP of all the classes. For OV-COCO, we follow previous work that only reports the $\rm{AP50^{b}}$ metric, which means the box mAP at IoU threshold $0.5$.

\noindent \textbf{Extension for Instance Segmentation.}
For OV-LVIS, instance segmentation results are needed for the evaluation process. Although DETR~\cite{carion2020end} and its follow-ups~\cite{zhu2020deformable,meng2021conditional} are developed for the object detection task, they can also be extended to the instance segmentation task. We follow DETR~\cite{carion2020end} to add an external class-agnostic segmentation head to solve the instance segmentation task. The segmentation head employs the fully convolutional network~(FCN~\cite{long2015fully}) structure, which takes features extracted from the Transformer decoder as input and produces segmentation masks.

\noindent \textbf{Implementation Details.}
Our model is based on Deformable DETR~\cite{zhu2020deformable}.
Following ViLD~\cite{gu2021open}, we also use the open-source CLIP model~\cite{radford2021learning} based on ViT-B/32 for extracting text and image embeddings.
Please refer to our supplementary material for more training details.

\subsection{Ablation Studies}\label{sec:ablation}

\begin{table*}[t]
\centering
\footnotesize
\setlength{\tabcolsep}{1pt}
\begin{minipage}{.53 \linewidth}
\centering
\setlength{\tabcolsep}{1pt}
\begin{tabular}{ll cccc}
\toprule
\rowNumber{\#} & Method & $\rm{AP^{m}}$ & $\rm{AP^{m}_{novel}}$ & $\rm{AP^{m}_{c}}$ & $\rm{AP^{m}_{f}}$ \\
\cmidrule(r){1-1}
\cmidrule(r){2-2}
\cmidrule(r){3-6}
\rowNumber{1} & \footnotesize{Mask R-CNN\dag}  & 22.5 & 0.0 & 22.6 & 32.4 \\
\rowNumber{2} & \footnotesize{Def DETR} & 22.4 & 0.0 & 22.4 & 32.0 \\
\bottomrule
\end{tabular}
\end{minipage}
\begin{minipage}{.44 \linewidth}
\caption{
\footnotesize{
\textbf{Mask R-CNN and Def DETR on OV-LVIS, both trained on base classes}.
$\dag$: copied from ViLD~\cite{gu2021open}.}
}
\label{tab:abla_arch}
\end{minipage}
\end{table*}

\begin{table*}[t]
\begin{minipage}{.48 \linewidth}
\centering
\setlength{\tabcolsep}{3pt}
\begin{tabular}{llcc cccc}
\toprule
\rowNumber{\#} & P & M & $\rm{AP^{m}}$ & $\rm{AP^{m}_{novel}}$ & $\rm{AP^{m}_{c}}$ & $\rm{AP^{m}_{f}}$ \\
\cmidrule(r){1-1}
\cmidrule(r){2-2}
\cmidrule(r){3-3}
\cmidrule(r){4-7}
\rowNumber{1} & ~ & ~ & 24.2 & 9.5 & 23.2 & 31.7 \\
\rowNumber{2} & \cmark & & 19.9 & 6.3 & 17.4 & 28.6 \\
\rowcolor{LavenderBlush} \rowNumber{3} & \cmark & \cmark & \textbf{26.6} & \textbf{17.4} & \textbf{25.0} & \textbf{32.5} \\
\bottomrule
\end{tabular}
\end{minipage}
\begin{minipage}{.44 \linewidth}
\footnotesize
\caption{
\footnotesize{\textbf{Ablation study} on using object proposals~(P) and our conditional binary matching mechanism~(M).}
}
\label{tab:abla_components}
\end{minipage}
\end{table*}

We conduct ablation study on OV-LVIS to evaluate the main components in our approach.

\noindent \textbf{The Architecture Difference.} Previous works such as ViLD~\cite{gu2021open} are based on the RPN-based Mask R-CNN~\cite{he2017mask}, while our work is based on the Transformer-based detector Deformable DETR~\cite{zhu2020deformable}. We first study the difference of these two detectors on the open-vocabulary setting trained with \emph{base} classes only. As shown in Table~\ref{tab:abla_arch} row(\rowNumber{1}-\rowNumber{2}), we observe that Mask R-CNN performs a slightly better than Deformable DETR~\cite{zhu2020deformable}.
This gap is small, indicating that we have a fair starting point compared to ViLD~\cite{gu2021open}.

\noindent \textbf{Object Proposals.}
We then replace Deformable DETR's classifier layer as text embedding provided by CLIP and trained with \emph{base} classes only.
This step is similar to the previous ViLD-text~\cite{gu2021open} method.
Results is presented in Table~\ref{tab:abla_components} row~\rowNumber{1}. We observe that the $\rm{AP^{m}_{novel}}$ metric improved from $0.0$ to $9.5$.
To further improve the $\rm{AP^{m}_{novel}}$ metric, we add the object proposals that may contain the region of \emph{novel} classes into the training stage.
Because we do not know the category id of these object proposals, we observe that the label assignment of these object proposals is inaccurate and will decrease the $\rm{AP^{m}_{novel}}$ performance from $9.5$ to $6.3$.

\noindent \textbf{Conditional Binary Matching.}
Now we replace DETR's default close-set labeling assignment as our proposed conditional binary matching. 
The comparison results between Table~\ref{tab:abla_components} row~\rowNumber{2-3} shows that our binary matching strategy can better leverage the knowledge from object proposals and improve the $\rm{AP^{m}_{novel}}$ from $9.5$ to $17.4$.
Such a large improvement shows that the proposed conditional matching is essential when applying the DETR-series detector for the open-vocabulary setting.

\subsection{Results on Open-vocabulary Benchmarks}
\begin{table}[t]
  \centering
  \caption{
  \footnotesize{
  \textbf{Main results on OV-LVIS and OV-COCO}.
  For OV-LVIS (w/ 886 base classes and 317 novel classes), we report mask mAP and a breakdown on novel (rare), common, and frequent classes. For OV-COCO (w/ 48 base classes and 17 novel classes), we report bounding box mAP at IoU threshold 0.5. \dag: zero-shot methods that do not use captions or image-text pairs. \ddag: ensemble model.
  }
  }
  \label{tab:main_results}
  \footnotesize
  \setlength{\tabcolsep}{3pt}
  \scalebox{1.00}{
   \begin{tabular}{l l cccc ccc}
    \toprule
    \multirow{2}{*}{\rowNumber{\#}} & \multirow{2}{*}{Method} & \multicolumn{4}{c}{\textbf{OV-LVIS}} & \multicolumn{3}{c}{\textbf{OV-COCO}} \\
    \cmidrule(r){3-6}
    \cmidrule(r){7-9}
     & & $\rm{AP^{m}}$ & $\rm{AP^{m}_{novel}}$ & $\rm{AP^{m}_{c}}$ & $\rm{AP^{m}_{f}}$ & $\rm{AP50^{b}}$ & $\rm{AP50^{b}_{novel}}$ & $\rm{AP50^{b}_{base}}$ \\
    \cmidrule(r){1-1}
    \cmidrule(r){2-2}
    \cmidrule(r){3-6}
    \cmidrule(r){7-9}
    \rowNumber{1} & SB~\cite{bansal2018zero}\dag & - & - & - & - & 24.9 & 0.3 & 29.2 \\
    \rowNumber{2} & DELO~\cite{zhu2020don}\dag & - & - & - & -  & 13.0 & 3.1 & 13.8 \\
    \rowNumber{3} & PL~\cite{rahman2021improved}\dag & - & - & - & -  & 27.9 & 4.1 & 35.9 \\
    \midrule
    \rowNumber{4} & OVR-CNN~\cite{zareian2021open} & - & - & - & -  & 46.0 & 22.8 & 39.9 \\
    \rowNumber{5} & ViLD-text~\cite{gu2021open} & 24.9 & 10.1 & 23.9 & \textbf{32.5} & 49.3 & 5.9 & \textbf{61.8} \\
    \rowNumber{6} & ViLD~\cite{gu2021open} & 22.5 & 16.1 & 20.0 & 28.3 & 51.3 & 27.6 & 59.5 \\
    \rowNumber{7} & ViLD-ens.~\cite{gu2021open}\ddag & 25.5 & 16.6 & 24.6 & 30.3 & - & - & -  \\
    \midrule
    \rowcolor{LavenderBlush} \rowNumber{8} & \methodname & \textbf{26.6} & \textbf{17.4} & \textbf{25.0} & \textbf{32.5} & \textbf{52.7} & \textbf{29.4} & 61.0 \\
    \rowcolor{LavenderBlush} & (ours vs. \#\rowNumber{6}) & \small{\better{4.1}} & \small{\better{1.3}} & \small{\better{5.0}} & \small{\better{4.2}} & \small{\better{1.4}} & \small{\better{1.8}} & \small{\better{1.5}} \\
    \bottomrule
    \end{tabular}
    }
\end{table}

Table~\ref{tab:main_results} summarizes our results.
We compare our method with SOTA open-vocabulary detection methods including: (1) OVR-CNN~\cite{zareian2021open}~(see Table~\ref{tab:main_results} row \rowNumber{4}). It pre-trains the detector's projecting layer on image-caption pairs using contrastive loss and then fine-tunes on the object detection task; (2) Variants of ViLD~\cite{gu2021open} such as ViLD-text and ViLD-ensemble~(see Table~\ref{tab:main_results} rows \rowNumber{5-7}). ViLD is the first study that uses CLIP embeddings~\cite{radford2021learning} for open-vocabulary detection. Compared with ViLD-text, ViLD uses knowledge distillation from the CLIP visual backbone, improves $\rm{AP_{novel}}$ at the cost of hurting $\rm{AP_{base}}$. ViLD-ens. combines the two models and shows improvements for both metrics. Such an ensemble-based method also brings extra time and memory cost.

For completeness, we also list the results of some previous zero-shot methods such as SB~\cite{bansal2018zero}, DELO~\cite{zhu2020don} and PL~\cite{rahman2021improved} in Table~\ref{tab:main_results} rows~\rowNumber{1-3}.
On \textit{OV-LVIS} benchmark, \methodname improves the previous SOTA ViLD by 4.1 on $\rm{AP^{m}}$ and 1.3 on $\rm{AP^{m}_{novel}}$.
Compared with ViLD, our method will not affect the performance of \emph{base} classes when improve the \emph{novel} classes.
Even compared with the ensemble result of ViLD-ensemble, \methodname still boosts the performance by 1.5, 0.8, 1.0 and 2.2, respectively (\%).
Noted that \methodname only uses a single model and does not leverage any ensemble-based technique.
On \textit{OV-COCO} benchmark, \methodname improves the baseline and outperforms OVR-CNN~\cite{zareian2021open} by a large margin, notably, the 6.6 mAP improvements on \emph{novel} classes.
Compared with ViLD~\cite{gu2021open}, \methodname still achieves 1.4 mAP gains on all the classs and 1.8 mAP gains on \emph{novel} classes.
In summary, it is observed that that \methodname achieves superior performance across different datasets compared with different methods.

\subsection{Generalization Ability of \methodname.}
\begin{table}[t]
  \centering
  \caption{ \footnotesize{
  \textbf{Generalization to other datasets}. We evaluate \methodname trained on LVIS when transferred to other datasets such as PASCAL VOC 2007 test set and COCO validation set by simply replacing the text embeddings. The experimental setting is the same as that of ViLD~\cite{gu2021open}. We observe that \methodname achieves better generalization performance than ViLD~\cite{gu2021open}.}
}
  \label{tab:transfer}
  \footnotesize
  \setlength{\tabcolsep}{3pt}
  \scalebox{1.05}{
   \begin{tabular}{ll cc ccc}
    \toprule
    \multirow{2}{*}{\rowNumber{\#}} & \multirow{2}{*}{Method} & \multicolumn{2}{c}{\textbf{Pascal VOC}} & \multicolumn{3}{c}{\textbf{COCO}} \\
    \cmidrule(r){3-4}
    \cmidrule(r){5-7}
     & & $\rm{AP_{50}^{b}}$ & $\rm{AP_{75}^{b}}$ & $\rm{AP}^{b}$ & $\rm{AP_{50}^{b}}$ & $\rm{AP_{75}^{b}}$ \\
    \cmidrule(r){1-1}
    \cmidrule(r){2-2}
    \cmidrule(r){3-4}
    \cmidrule(r){5-7}
    \rowNumber{1} & ViLD-text~\cite{gu2021open} & 40.5 & 31.6 & 28.8 & 43.4 & 31.4  \\
    \rowNumber{2} & ViLD~\cite{gu2021open} & 72.2 & 56.7 & 36.6 & 55.6 & 39.8 \\
    \midrule
    \rowcolor{LavenderBlush} \rowNumber{3} & \methodname & \textbf{76.1} & \textbf{59.3} & \textbf{38.1} & \textbf{58.4} & \textbf{41.1} \\
    \rowcolor{LavenderBlush} & (ours vs \rowNumber{\#2} ) & \better{3.9} & \better{2.6} & \better{1.5} & \better{2.8} & \better{1.3} \\
    \bottomrule
    \end{tabular}
    }
\end{table}
We follow ViLD~\cite{gu2021open} to test the generalization ability of \methodname by training the model on LVIS~\cite{gupta2019lvis} dataset and evaluated on PASCAL VOC~\cite{everingham2010pascal} and COCO~\cite{lin2014microsoft}.
We keep the same implementation details with  ViLD~\cite{gu2021open}.
We switch the text embeddings of the category names from the source dataset to new datasets.
The text embeddings of new classes are used as conditional inputs during the inference phase.
As shown in Table~\ref{tab:transfer}, we observe that \methodname achieves better transfer performance than ViLD.
The experimental results show that the model trained by our conditional-based mechanism has transferability to other domains.

\subsection{Qualitative Results}
\begin{figure*}[t]
\centering
   \includegraphics[width=0.99\linewidth]{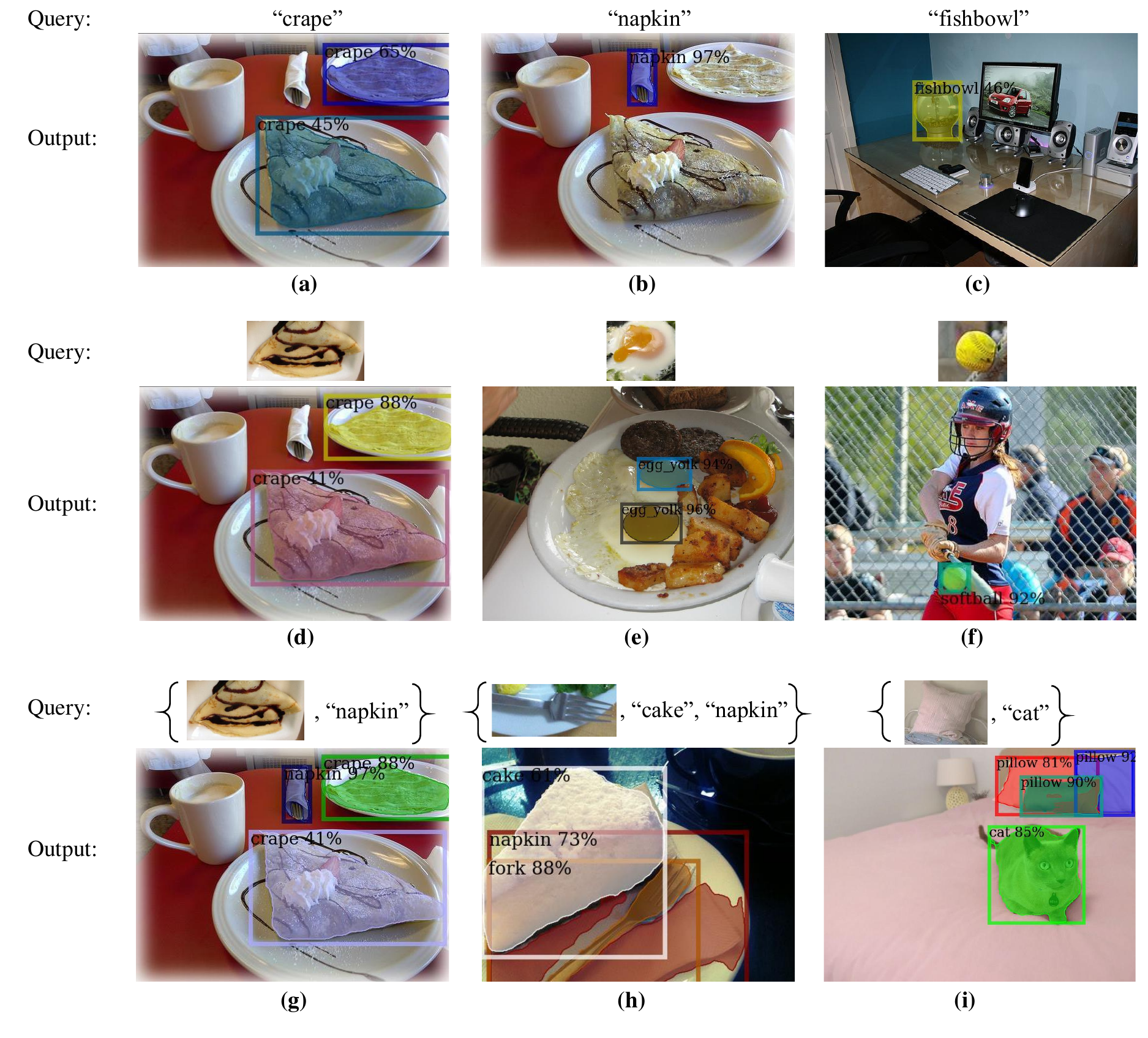}
   \caption{\textbf{Qualitative results on LVIS.}
   \methodname can precisely detect and segment novel objects~(\eg, `crape', `fishbowl', `softball') given the conditional text query~(top) or conditional image query~(middle) or a mixture of them~(bottom).
   }
   \label{fig:main_qual}
\end{figure*}

We visualize OV-DETR's detection and segmentation results in Fig.~\ref{fig:main_qual}. The results based on conditional text queries, conditional image queries, and a mixture of conditional text and image queries are shown in the top, middle and bottom row, respectively. Overall, our OV-DETR can accurately localize and precisely segment out the target objects from novel classes despite no annotations of these classes during training. It is worth noting that the conditional image queries, such as ``crape'' in (d) and ``fork'' in (h), appear drastically different from those in the target images but OV-DETR can still robustly detect them.

\subsection{Inference Time Analysis}
\begin{table}[t]
\centering
\setlength{\tabcolsep}{3pt}
\begin{minipage}{.6 \linewidth}
\footnotesize
\centering
\setlength{\tabcolsep}{3pt}
\begin{tabular}{ll cc}
        \toprule
         \rowNumber{\#} & Method & COCO & LVIS \\
         \cmidrule(r){1-1}
         \cmidrule(r){2-2}
         \cmidrule(r){3-3}
         \cmidrule(r){4-4}
         \rowNumber{1} & Def DETR & 0.31 & 1.49  \\
         \rowNumber{2} & Ours & 0.72 & 23.84 \\
         \rowNumber{3} & Ours~(optimized) & 0.63 & 9.57 \\
         \rowcolor{LavenderBlush} & (vs \#\rowNumber{2}) & \better{$\downarrow 12.5\%$} & \better{$\downarrow 59.9\%$} \\
         \bottomrule
    \end{tabular}
\end{minipage}
\hspace{+4pt}
\begin{minipage}{.35 \linewidth}
\centering
\caption{
\footnotesize{
Comparison of the inference time~(second per iteration) between Deformable DETR~\cite{zhu2020deformable} and our~\methodname before/after optimization on LVIS and COCO.
}}
\label{tab:inference_time}
\end{minipage}
\end{table}

OV-DETR exhibits great potential in open-vocabulary detection but is by no means a perfect detector. The biggest limitation of OV-DETR is that the inference speed is slow when the number of classes to detect is huge like 1,203 on LVIS~\cite{gu2021open}. This problem is caused by the conditional design that requires multiple forward passes in the Transformer decoder (depending on the number of classes). 

We show a detailed comparison on the inference time between Deformable DETR and OV-DETR in Table~\ref{tab:inference_time}. Without using any tricks, the vanilla OV-DETR (\#\rowNumber{2}), i.e., using a single forward pass for each class, is about 2$\times$ slower than Deformable DETR (\#\rowNumber{1}) on COCO (w/ 80 classes) while 16$\times$ slower on LVIS (w/ 1,203 classes).
As discussed in Sec.~\ref{subsec:conditional_matching} and shown in Fig.~\ref{fig:conditional_query}(b), we optimize the speed by forwarding multiple conditional queries to the Transformer decoder in parallel, which reduces the inference time by 12.5\% on COCO and nearly 60\% on LVIS (see \#\rowNumber{3} in Table~\ref{tab:inference_time}). Still, there is much room for improvement.

It is worth noting that such a slow inference problem is not unique to our approach---most instance-conditional models would have the same issue~\cite{li2017person}, which is the common price to pay in exchange for better performance. The computation bottleneck of our method lies in the computation of the Transformer decoder in Eq.~(\ref{eq:detr_set}). A potential solution is to design more efficient attention modules~\cite{tay2020sparse,wang2020linformer}, which we leave as future work.
In human-computer interaction where users already have target object(s) in mind, \eg, a missing luggage or a specific type of logo, the conditional input is fixed and low in number, thus the inference time is negligible.
\section{Conclusion}
Open-vocabulary detection is known to be a challenging problem due to the lack of training data for unseen classes. Recent advances in large language models have offered a new perspective for designing open-vocabulary detectors. In this work, we show how an end-to-end Transformer-based detector can be turned into an open-vocabulary detector based on conditional matching and with the help of pre-trained vision-language models. The results show that, despite having a simplified training pipeline, our open-vocabulary detector based on Transformer significantly outperforms current state of the arts that are all based on two-stage detectors. We hope our approach and the findings presented in the paper can inspire more future work on the design of efficient open-vocabulary detectors.

\noindent \textbf{Acknowledgment}
This study is supported under the RIE2020 Industry Alignment Fund Industry Collaboration Projects (IAF-ICP) Funding Initiative, as well as cash and in-kind contribution from the industry partner(s). It is also partly supported by the NTU NAP grant and Singapore MOE AcRF Tier 2 (MOE-T2EP20120-0001). This work was supported by SenseTime SenseCore AI Infrastructure-AIDC. 

\clearpage
%
%
\bibliographystyle{splncs04}
\bibliography{egbib}
\clearpage
\appendix
\begin{center}
 \textbf{\large Supplementary Material}
\end{center}

\section{More Qualitative Results of \methodname}
\begin{figure*}[t]
\begin{minipage}{1.0 \linewidth}
\centering
   \includegraphics[width=0.99\linewidth]{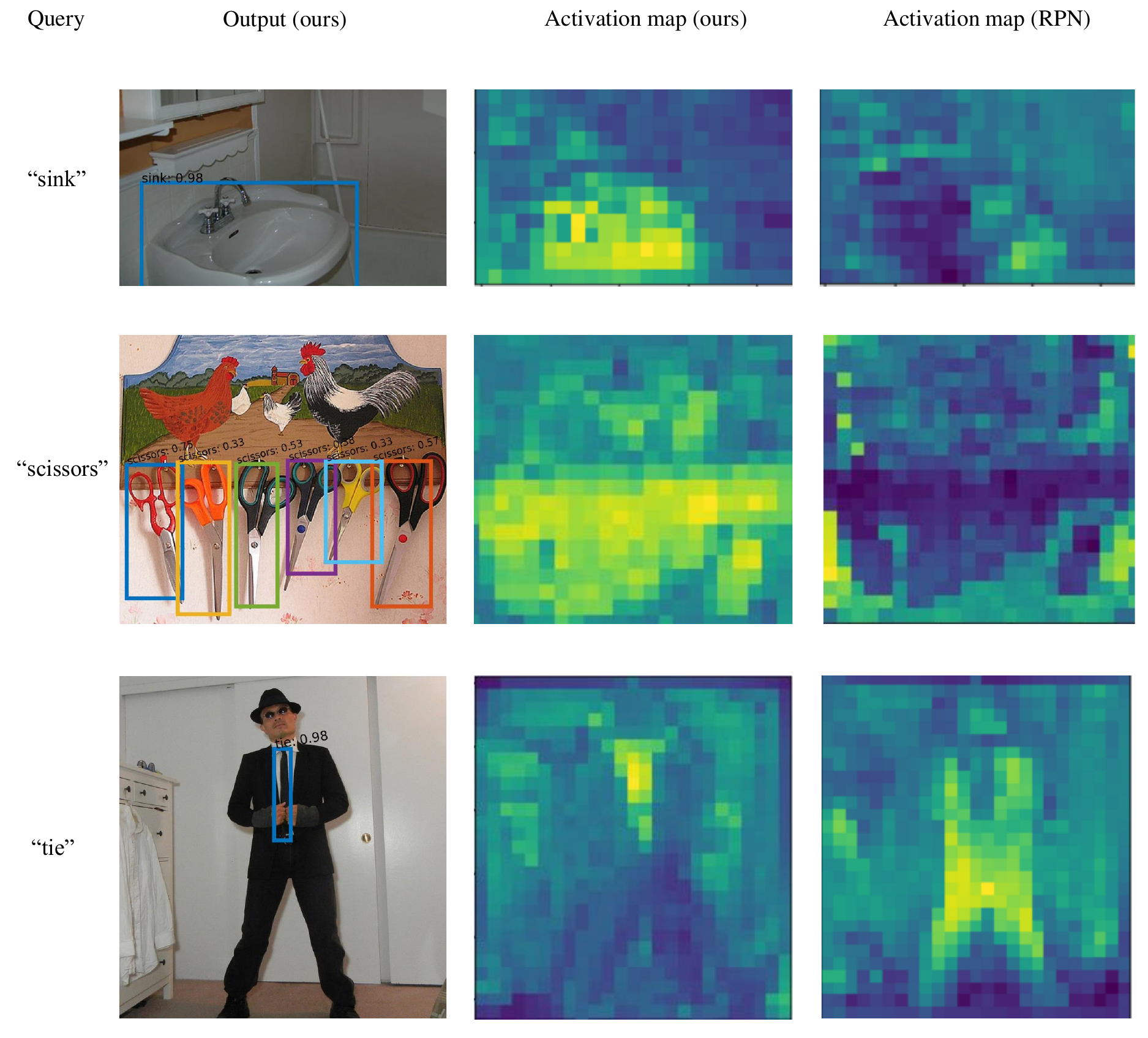}
   \caption{
   \textbf{Qualitative results on Open-Vocabulary COCO setting.}
   We visualize the prediction results of \methodname on \emph{novel} classes.
   We also provide the comparison of activation maps between ours and the RPN network.
   \newline
   }
   \label{fig:appendix_qual}
\end{minipage}
\begin{minipage}{1.0 \linewidth}
\includegraphics[width=0.99\linewidth]{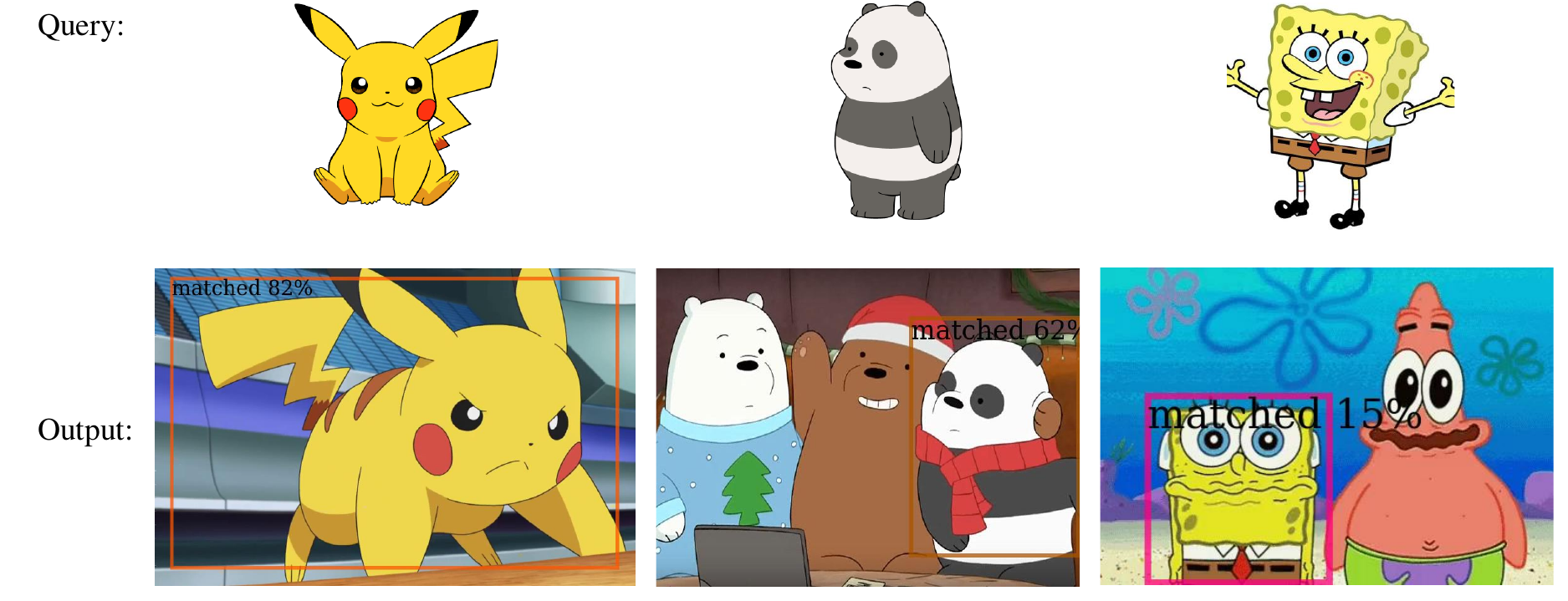}
   \caption{
   \textbf{Qualitative results on anime characters.} These images are collected from web. We use the model trained on LVIS dataset to check the matchability with the given conditional image queries.
   }
   \label{fig:appendix_qual_web}
\end{minipage}
\end{figure*}

\noindent \textbf{Open-Vocabulary COCO.}
We provide more qualitative results of \methodname on Open-Vocabulary COCO setting~(Fig.~\ref{fig:appendix_qual}).
We visualize the detection results on \emph{novel} classes and the activation maps of \methodname and Region Proposal Network~(RPN)~\cite{ren2015faster} used by ViLD~\cite{gu2021open}, which further validate the motivation from the main paper: \methodname has higher activation values on objects of \emph{novel} classes than RPN.

\noindent \textbf{Web Images.}
To verify the generalization ability, we also provide the qualitative results of anime characters in Fig.~\ref{fig:appendix_qual_web}.
Although these characters are not provided during training, \methodname can successfully detect the regions matched with the conditional image queries.

\noindent \textbf{Failure Cases.}
Fig.~\ref{fig:appendix_qual_failture} shows some failure cases of \methodname.
We notice that detecting small or occluded objects with conditional image query is hard.
Our method is not robust to the unrelated out-of-distribution text queries.
We will address these shortcomings in further research.


\section{Discussion of Object Proposals}

\begin{figure}[t]
\begin{minipage}{1.0 \linewidth}
\includegraphics[width=0.99\linewidth]{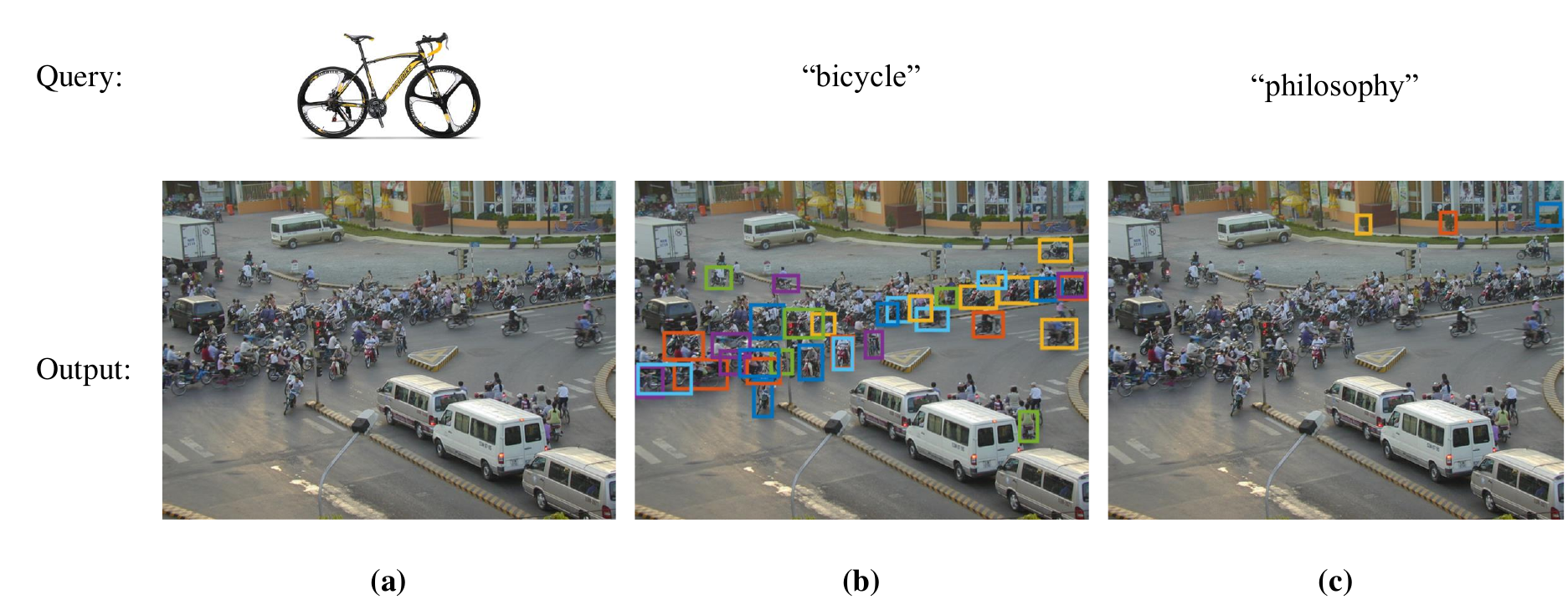}
   \caption{
   \textbf{Failure cases of \methodname.}
   These images are collected from the web. We use the model trained on LVIS dataset to check the matchability with the given conditional image or text queries.
   \textbf{(a):} \methodname fails to detect these small and occluded objects with the conditional image query~(``bicycle''). But as shown in \textbf{(b)}, this issue can be solved to some extent by using text query.
   \textbf{(c):} Given the unrelated text queries~(\eg, ``philosophy''), \methodname will predict wrong false-positive detection results.
   \newline
   }
   \label{fig:appendix_qual_failture}
\end{minipage}
\begin{minipage}{1.0 \linewidth}
\centering
   \includegraphics[width=0.99\linewidth]{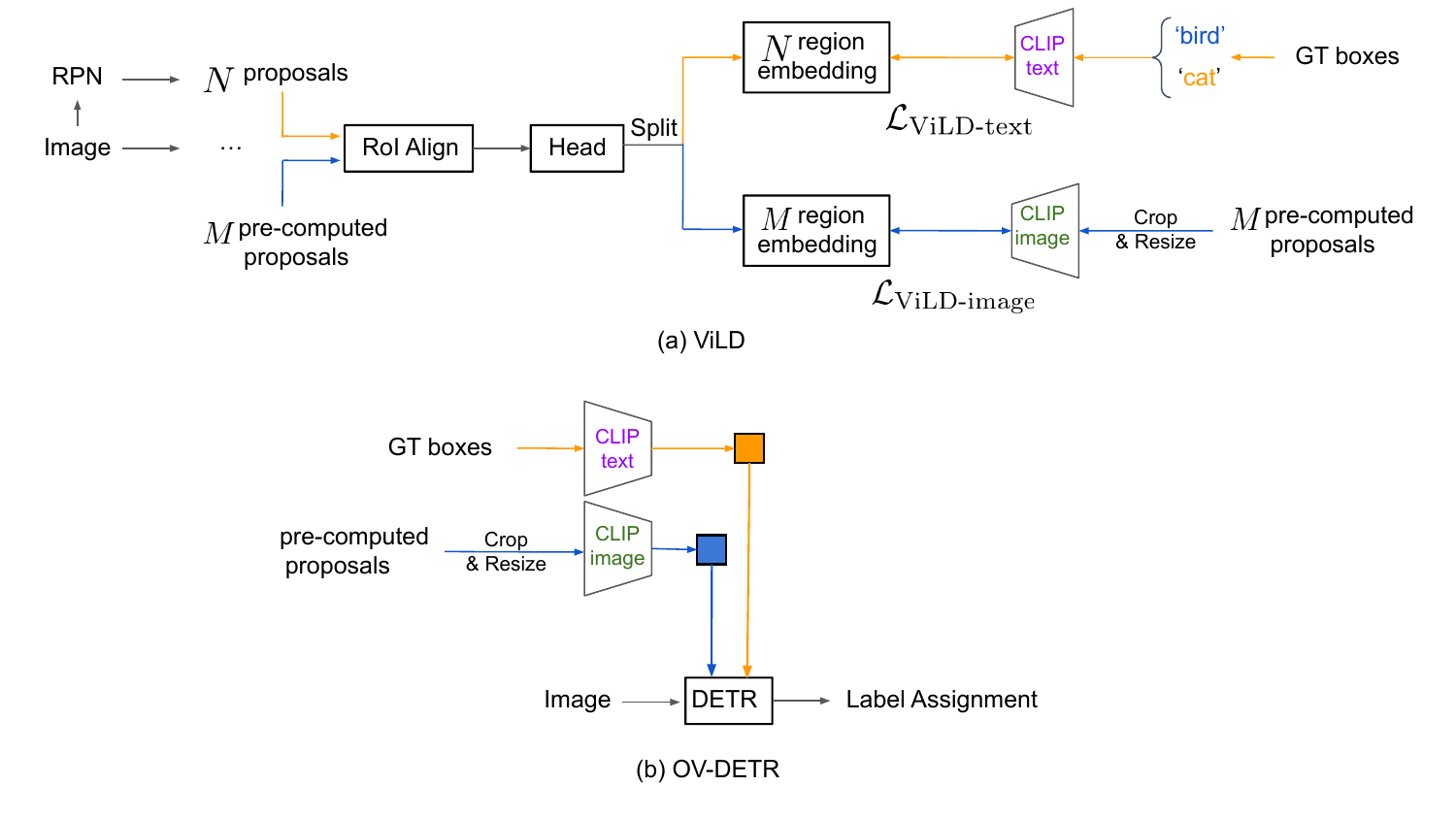}
   \caption{
   The comparison of ViLD and our \methodname of utilizing pre-computed object proposals.
   \textbf{(a):} ViLD leverages a distillation loss between the predicted region embeddings and pre-computed object proposals.
   \textbf{(b):} We use pre-computed object proposals to generate the conditional image query.
   }
   \label{fig:appendix_proposal}
\end{minipage}
\end{figure}

In previous work ViLD~\cite{gu2021open}, class-agnostic object proposals are leveraged to transfer the knowledge from CLIP image encoder to the detector.
As shown in Fig.~\ref{fig:appendix_proposal}~(a), ViLD first trains a RPN network on \emph{base} classes to get $M$ pre-computed proposals. These object proposals may contain objects of \emph{novel} classes and are essential for training ViLD model. For these $M$ proposals, $M$ predicted region embeddings and the corresponding ground-truth embeddings are computed by a Mask R-CNN detector and a CLIP image encoder respectively.
Then, a knowledge distillation loss is applied for the predicted region embeddings and the ground-truth embeddings.

We follow the steps of ViLD, pre-training a detector with the \emph{base} classes to predict object proposals that may cover the \emph{novel} classes.
The only difference is that we use different architectures (Def-DETR vs. the RPN network in ViLD).
Despite of building upon different architectures, the generated object proposals have similar high top-300 averaged recall (AR@300) for \emph{novel} categories (48.3 for ViLD and 47.6 for ours).
Fig.~\ref{fig:appendix_proposal}~(b) shows that compared with ViLD, we use the object proposals in a different way.
Since these object proposals are class-agnostic, they cannot be applied on DETR's matching algorithm.
In \methodname, we treat the image embeddings extracted by the object proposals as the conditional image query.
The expected predictions of \methodname are the `matched' regions of an input image given a conditional image query.

\section{Importance of $\mathcal{L}_{embed}$} 
\begin{table}[t]
\centering
\setlength{\tabcolsep}{3pt}
\begin{minipage}{.99 \linewidth}
\centering
\caption{
\footnotesize{Importance of $\mathcal{L}_{\text{embed}}$ on LVIS.}
}
\label{tab:appendix_embedding}
\scalebox{1.0}{
\setlength{\tabcolsep}{3pt}
\begin{tabular}{lc cccc}
\toprule
\rowNumber{\#} & $\mathcal{L}_{\text{embed}}$ & $\rm{AP^{m}}$ & $\rm{AP^{m}_{novel}}$ & $\rm{AP^{m}_{c}}$ & $\rm{AP^{m}_{f}}$  \\
\cmidrule(r){1-1}
\cmidrule(r){2-2}
\cmidrule(r){3-6}
\rowNumber{1} & \xmark     & 24.9 & 14.4 & 23.2 & 31.3 \\
\rowcolor{LavenderBlush} \rowNumber{2} & \checkmark & \textbf{26.6} & \textbf{17.4} & \textbf{25.0} & \textbf{32.5} \\
\bottomrule
\end{tabular}}
\end{minipage}
\vspace{-12pt}
\end{table}

In \methodname, we introduce an embedding reconstruction head to predict the conditional input embedding $\vz^\text{text}$ or $\vz^\text{image}$, and this reconstruction head is optimized by the loss $\mathcal{L}_{embed}$.
The results in Table~\ref{tab:appendix_embedding} show the efficacy of $\mathcal{L}_{embed}$.

\begin{table}[t]
    \footnotesize
    \centering
    \caption{
    \footnotesize{\textbf{Ablation study} on $N$ (the number of object queries) and $R$ (the number of copies).}
    }
    \label{tab:abla_repeats}
    \scalebox{1.0}{
    \setlength{\tabcolsep}{3pt}
    \begin{tabular}{lll cccc}
        \toprule
         \rowNumber{\#} & $N$ & $R$ & $\rm{AP^{m}}$ & $\rm{AP^{m}_{novel}}$ & $\rm{AP^{m}_{c}}$ & $\rm{AP^{m}_{f}}$ \\
         \cmidrule(r){1-1}
         \cmidrule(r){2-3}
         \cmidrule(r){4-7}
        \rowNumber{1} & 100 & 1 & 22.0 & 10.6 & 20.9 & 28.2 \\
        \rowNumber{2} & 100 & 3 & 25.7 & 13.6 & \textbf{25.0} & 31.9 \\
        \rowNumber{3} & 100 & 9 & 24.3 & 11.9 & 22.9 & 31.3 \\
        \rowNumber{4} & 300 & 1 & 24.2 & 12.3 & 22.8 & 30.9 \\
        \rowcolor{LavenderBlush} \rowNumber{5} & 300 & 3 & \textbf{26.6} & \textbf{17.4} & \textbf{25.0} & \textbf{32.5} \\
        \bottomrule
    \end{tabular}
    }
\end{table}

\section{Importance of Multiple Queries for Training}~\label{sec:abla_repeat_query}
\noindent Recap that we propose to ``clone'' query features in Section~3.2 (also see Fig~4) of the main paper.
We examine different choices of the two hyper-parameters $N$ and $R$, and show results in Table~\ref{tab:abla_repeats}.
When $N=100$, we find that coping queries (from $R=1$ to 3) improves the $\rm{AP^{m}_{novel}}$ from $10.6$ to $13.6$, and a slight degradation when $R=9$ partially due to the limited optimization capacity.
When $N=300$, we observe that coping queries (from $R=1$ to 3) is also beneficial.
However, we will face the out-of-memory issue on GPU when $N=300$ and $R>3$.
Overall, we find the combination of $N=300$ queries and repetition of $R=3$ times serves as the optimal solution.

\section{More Implementation Details}
\noindent \textbf{Hyper-Parameters.}
All models are trained on 8 Tesla V100 GPUs.
We use the ResNet50-C4 backbone as our default choice.
We keep most of the hyper-parameters the same with previous works~\cite{gu2021open,zhu2020deformable}.
For loss functions, we set the weighting parameters $\mathcal{L}_{\text{BCE}}=3.0$, $\mathcal{L}_{\text{L1}}=5.0$, $\mathcal{L}_{\text{GIoU}}=2.0$ and $\mathcal{L}_{\text{embed}}=1.0$.
The input resolution of the CLIP model is set to 224x224, and the temperature $\tau$ of the CLIP model is set to 0.01.

\noindent \textbf{Text Prompts.}
Prompt tuning is a critical step when transferring pre-trained language models to downstream computer vision tasks~\cite{zhou2022coop,zhou2022cocoop,zhang2022neural}. We follow the same process as in ViLD~\cite{gu2021open} to construct the text prompts. Specifically, for each class we feed the textual name wrapped in 63 different prompt templates (e.g., \texttt{`there is a \{class name\} in the photo'}) to CLIP's text encoder, and then average the 63 text embeddings, which is known as prompt ensembling~\cite{radford2021learning}.

\end{document}